\documentclass[journal]{vgtc}                % final (journal style)
\ifpdf%                                % if we use pdflatex
  \pdfoutput=1\relax                   % create PDFs from pdfLaTeX
  \usepackage{graphicx}                % allow us to embed graphics files
  \DeclareGraphicsExtensions{.pdf,.png,.jpg,.jpeg} % for pdflatex we expect .pdf, .png, or .jpg files
\else%                                 % else we use pure latex
  \usepackage{graphicx}                % allow us to embed graphics files
  \DeclareGraphicsExtensions{.eps}     % for pure latex we expect eps files
\fi%

\graphicspath{{figures/}{pictures/}{images/}{./}} % where to search for the images

\usepackage{microtype}                 % use micro-typography (slightly more compact, better to read)
\PassOptionsToPackage{warn}{textcomp}  % to address font issues with \textrightarrow
\usepackage{textcomp}                  % use better special symbols
\usepackage{mathptmx}                  % use matching math font
\usepackage{times}                     % we use Times as the main font
\usepackage{cite}                      % needed to automatically sort the references
\usepackage{tabu}                      % only used for the table example
\usepackage{booktabs}                  % only used for the table example
\usepackage{comment}
\usepackage{amsmath,amssymb} % define this before the line numbering.
\usepackage{color}
\usepackage{capt-of}
\usepackage{cuted}
\usepackage{xcolor}
\usepackage{hyperref}
\usepackage{wrapfig}
\usepackage{tabu}
\usepackage{tcolorbox}
\usepackage{scalerel}
\usepackage{comment}
\usepackage{amsmath,amssymb} % define this before the line numbering.
\usepackage{color}
\usepackage{capt-of}
\usepackage{cuted}
\usepackage{xcolor}
\usepackage{hyperref}
\usepackage{wrapfig}
\usepackage{tcolorbox}
\usepackage{enumitem}
\usepackage[nameinlink]{cleveref}

\definecolor{taskcolor}{HTML}{2196F3}
\newtcbox{\taskbox}{on line, colframe=taskcolor,colback=taskcolor!10!white, boxrule=0.5pt,arc=4pt,boxsep=0pt,left=2pt,right=2pt,top=2pt,bottom=2pt}
\definecolor{technicalcolor}{HTML}{f44336}
\newtcbox{\technicalbox}{on line, colframe=technicalcolor,colback=technicalcolor!10!white, boxrule=0.5pt,arc=4pt,boxsep=0pt,left=2pt,right=2pt,top=2pt,bottom=2pt}
\definecolor{ethicalcolor}{HTML}{FF9800}
\newtcbox{\ethicalbox}{on line, colframe=ethicalcolor,colback=ethicalcolor!10!white, boxrule=0.5pt,arc=4pt,boxsep=0pt,left=2pt,right=2pt,top=2pt,bottom=2pt}
\definecolor{guidecolor}{HTML}{8BC34A}
\newtcbox{\guidebox}{on line, colframe=guidecolor,colback=guidecolor!10!white, boxrule=0.5pt,arc=4pt,boxsep=0pt,left=2pt,right=2pt,top=2pt,bottom=2pt} 
\definecolor{figurecolor}{HTML}{9E9E9E}
\newtcbox{\figurebox}{on line, colframe=figurecolor,colback=figurecolor!10!white, boxrule=0.5pt,arc=4pt,boxsep=0pt,left=2pt,right=2pt,top=2pt,bottom=2pt,fontupper=\sffamily} 

\onlineid{0}

\vgtccategory{Research}
\vgtcpapertype{application}

\title{
Visual Identification of Problematic Bias\\in Large Label Spaces
}

\author{
Alex Bäuerle, Aybuke Gul Turker, Ken Burke, Osman Aka, Timo Ropinski, Christina Greer, and Mani Varadarajan % <-this % stops a space
}
\authorfooter{
A. Bäuerle and T. Ropinski were with the Visual Computing Group at Ulm University.
E-mail: see \url{https://a13x.io}.\\
Aybuke Gul Turker and Mani Varadarajan were with the TensorBoard group at Google Research.\\
Ken Burke, Osman Aka and Christina Greer were with the Fairness Infrastructure group at Google Research.
}

\shortauthortitle{Bäuerle \MakeLowercase{\textit{et al.}}: Visual Identification of Problematic Bias in Large Label Spaces}
\abstract{
While the need for well-trained, fair ML systems is increasing ever more, measuring fairness for modern models and datasets is becoming increasingly difficult as they grow at an unprecedented pace.
One key challenge in scaling common fairness metrics to such models and datasets is the requirement of exhaustive ground truth labeling, which cannot always be done.
Indeed, this often rules out the application of traditional analysis metrics and systems.
At the same time, ML-fairness assessments cannot be made algorithmically, as fairness is a highly subjective matter.
Thus, domain experts need to be able to extract and reason about bias throughout models and datasets to make informed decisions.
While visual analysis tools are of great help when investigating potential bias in DL models, none of the existing approaches have been designed for the specific tasks and challenges that arise in large label spaces.
Addressing the lack of visualization work in this area, we propose guidelines for designing visualizations for such large label spaces, considering both technical and ethical issues.
Our proposed visualization approach can be integrated into classical model and data pipelines, and we provide an implementation of our techniques open-sourced as a TensorBoard plug-in.
With our approach, different models and datasets for large label spaces can be systematically and visually analyzed and compared to make informed fairness assessments tackling problematic bias.
} % end of abstract


\CCScatlist{ % not used in journal version
 \CCScat{K.6.1}{Management of Computing and Information Systems}%
{Project and People Management}{Life Cycle};
 \CCScat{K.7.m}{The Computing Profession}{Miscellaneous}{Ethics}
}

\vgtcinsertpkg

\begin{document}
%-------------------------------------------------------------------------
\firstsection{Introduction\label{sec:introduction}}

\maketitle

\begin{figure}[tb]
    \centering
    \includegraphics[width=.8\linewidth]{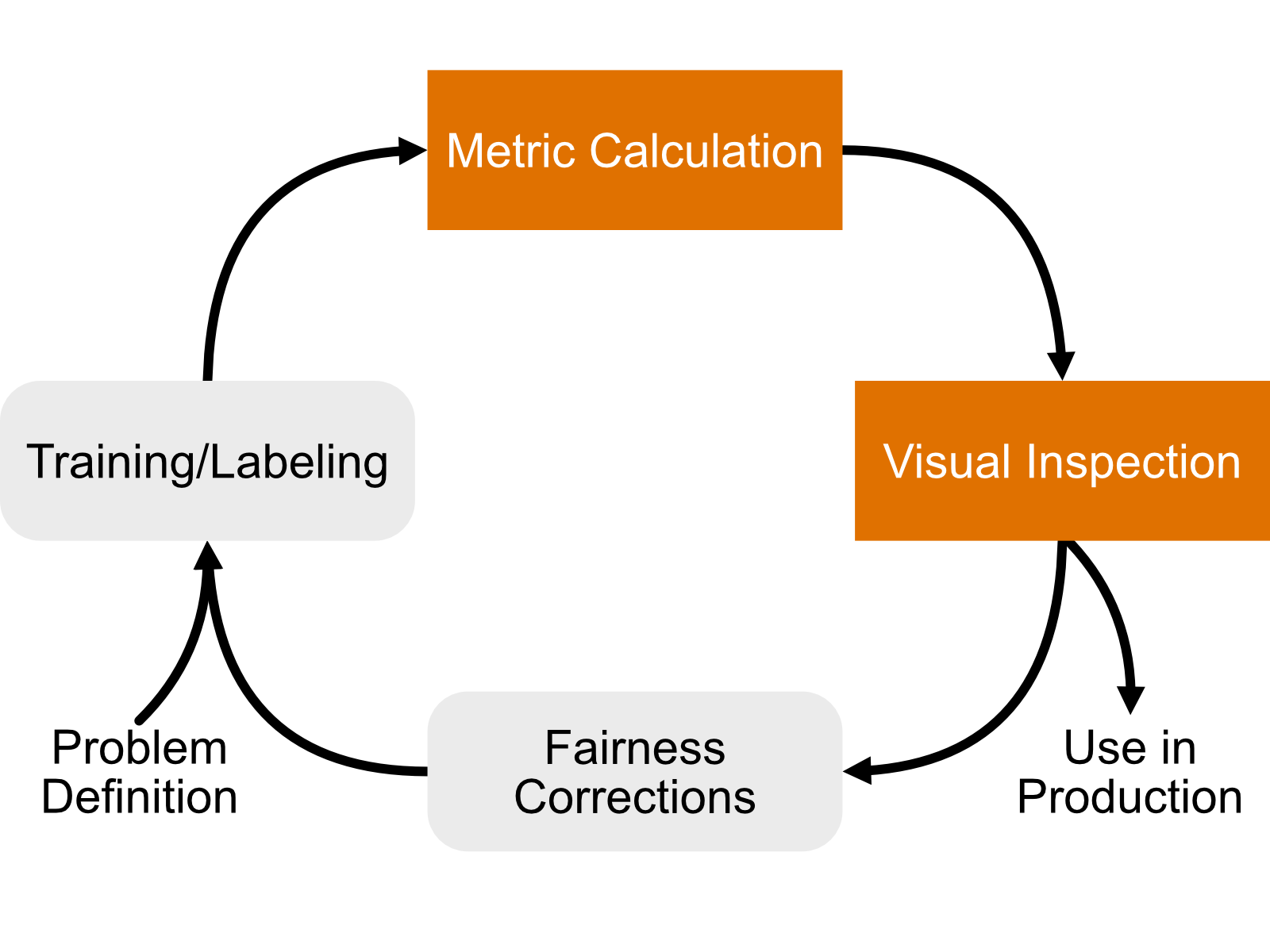}
    \caption{\label{fig:process} 
    Our approach is designed to be integrated into standard ML workflows for dataset acquisition and model training in large label spaces.
    Grey boxes represent steps that are already a part of many standard ML workflows, whereas orange rectangles are how we propose to enable a visual inspection of bias.
    Such a workflow begins with the labeling of data or training of a model.
    We then propose to use correlation metrics, to assess bias in these settings.
    Next, our proposed visual inspection design comes into play.
    Here decision makers can investigate bias, and assess whether it must be considered a problematic bias.
    Finally, either fairness corrections have to be made before reiterating on the process, or, when fairness concerns have already been alleviated, the model can be used in production.
    }
\end{figure}

The application areas of deep learning (DL) techniques broaden every day.
With this, these systems become an integral part of our daily lives, for instance in navigation, photos, and weather.
We even see the adoption of such techniques for critical domains, such as healthcare and finance.
Thus, fairness becomes more and more important~\cite{eu2020altai,jobin2019global}.
At the same time, when looking at the literature, the impact of algorithmic bias of DL models is  apparent~\cite{buolamwini2018gender,snow2018amazon,wilson2019predictive,hendricks2018women,stock2018convnets,obermeyer2019dissecting}.
The fact that most of the systems observed in the aforementioned works are deployed, operational, and public-facing, only makes this more alarming.
\\
As such, the need for well-trained, fair systems is increasing evermore.
However, measuring fairness for modern models and datasets is becoming increasingly difficult as they grow at an unprecedented pace.
While in 2008, the then introduced PASCAL VOC had 20 classes~\cite{everingham2015pascal}, it is nowadays common to operate on the full ImageNet set with over 20,000 categories~\cite{imagenet_cvpr09}.
In this paper, we focus on large label spaces (see~\autoref{subsec:definitions} for a definition of this and other terms), where datasets contain 1,000s of labels or more and multilabel classification models for such label spaces.
Problems of this scale can be seen in different datasets, such as the aforementioned ImageNet challenge~\cite{imagenet_cvpr09}, or the Open Images Dataset, which both contain about 20,000 categories~\cite{kuznetsova2018open}, as well as in modern applications, such as photo-tagging, text annotation, photo-search, and many more.
To be able to operate on these large label spaces, the amount of training data naturally also has to grow.
\\
One key challenge in scaling common fairness metrics to such models and datasets is the requirement of comprehensive ground truth labeling.
Here, classical approaches, such as gathering multiple labels for each data point, or even exhaustively labeling the dataset, are not feasible because of the sheer amount of data and labels.
For instance, obtaining exhaustive ground truth for OID would require about 20,000 (\# classes) times 10,000,000 labels (\# data points), which is not feasible to obtain.
Therefore, such datasets are, in most cases, sparsely labeled.
This renders classical fairness metrics, such as Statistical Parity~\cite{dwork2012fairness}, Equality of Opportunity~\cite{hardt2016equality}, and Predictive Parity~\cite{chouldechova2017fair} inapplicable, as they rely on an exhaustive ground truth in the evaluation dataset.
This problem clearly grows worse as the number of classes grows.
However, the benefits of growing the number of classes in order to make one's ontological mapping of the world more fine-grained are also clear.
To circumvent this problem, for large label spaces, correlation-based fairness metrics which rely minimally on ground truth label availability have been recently proposed~\cite{aka2021measuring}.
\\
While such correlation metrics open up a new way of investigating fairness for non-exhaustive ground truth data, both model and label fairness are still highly subjective and dependant on the application scenario.
Thus, following ethics guidelines, we argue that fairness issues, even with fairness metrics at hand, cannot be addressed by automated systems~\cite{jobin2019global,oecd2019assessment,eu2020altai}.
The fact that the Assessment List for Trustworthy Artificial Intelligence (ALTAI) which was set up by the European commission lists human oversight as the first requirement in their list~\cite{eu2020altai}, and that Jobin et. al.~\cite{jobin2019global} found that the most prominent principal among their surveyed guidelines is \emph{transparency}, followed by \emph{justice and fairness} underlines the importance of this aspect.
\\
To give an example, say a model trained to assign thousands of labels to images of human faces is skewed towards assigning the label~\emph{high income} with more likelihood to the \emph{male} direction of the sensitive attribute gender.
Whether this poses a problem depends on a lot of different factors, such as the intended application area, purpose, impact, and role of the model.
If model designers want to just straight up reflect the reality of nature to raise awareness, and this is how their data is distributed, they might accept such a biased model (i.e. just aiming for high prediction accuracy).
However, if they are concerned about the broader social impact, reducing pre-existing bias in the world, or reputation of their company, they likely want to change the behavior of their model in order to mitigate such bias.
This difference is what separates bias from problematic bias.
As such assessments cannot be made algorithmically, domain experts need to be able to extract and reason about bias to make informed decisions.
\\
Visualization, as shown in recent work~\cite{ahn2019fairsight,cabrera2019fairvis}, can streamline and speed up such investigations drastically.
While visual analysis tools are of great help when investigating potential bias in DL models, none of the existing approaches have been designed for the specific tasks and challenges that arise in large label spaces, where in contrast, decision makers still look over large spreadsheets with fairness metric calculations.
Thus, current fairness visualization approaches do not support the sheer amount of data points and labels, as well as these novel metrics, and the way they are investigated for large label spaces.
\\
Addressing this lack of visualization work in this area, we propose \guidebox{guidelines} for designing visualizations for such large label spaces, which we extract from the \taskbox{tasks} of decision makers in such problem domains, and challenges, both \technicalbox{technical} and \ethicalbox{ethical}, which arise when designing such fairness visualizations.
Our proposed visualization approach can be integrated into classical model and data pipelines as shown in~\autoref{fig:process}, and we provide an implementation of our techniques which we open-sourced as a TensorBoard plugin~\url{https://github.com/tensorflow/tensorboard/tree/master/tensorboard/plugins/npmi}.
Our contributions to this end are threefold:
\begin{enumerate}[label=C\arabic*]
    \item We identify tasks and challenges specific for identifying problematic bias in large label spaces.
    \item We propose visualization design guidelines that are aimed at tackling these challenges.
    \item We provide a public implementation and evaluation of these visualization guidelines.
\end{enumerate}

Our visualization design challenges and guidelines are directly informed by the tasks of practitioners in the field, with which we collaborated closely throughout this research.
Our approach has been tested and integrated in a variety of real-world applications by a range of practitioners and proved to be helpful in the practice of spotting potential bias in both labeling processes and model predictions.

%-------------------------------------------------------------------------
\section{Related Work}

While there have been many visualization approaches targeted towards large datansets, such as active learning~\cite{heimerl2012visual,hoferlin2012inter,kucher2017active}, interactive labeling~\cite{bernard2018vial,bernard2017comparing}, labeling process improvements~\cite{liu2018interactive,chang2017revolt}, and data cleaning~\cite{xiang2019interactive,bauerle2020classifier}, they are all used during the labeling process.
Fairness analysis, on the contrary, happens either after dataset collection has been completed (data fairness), or even after model training has been finished (model fairness).
Thus, in the following, our focus is on visualization techniques for bias mitigation.

The domain of ML-fairness related visualization work has seen significant attention in the last few years~\cite{wang2020visual,law2020impact,cheng2020dece}, which can be observed by the growing research interest~\cite{chatzimparmpas2020state,yuan2020survey}.
To support this need for visualization-based ML-fairness methods, an interview study~\cite{law2020designing} with domain experts found that for resolving fairness problems, people want to have automatic tools to detect bias.
However, they are reliant on visualization to investigate bias as this is a highly subjective manner, which has also been observed by Zhang et. al.~\cite{zhang2020joint}.
The following paragraphs will cover a representative subset of the work in this domain.
\\
Themis-ML~\cite{bantilan2018themis} measures fairness problems based on classical fairness metrics and integrates mitigation methods into the modeling pipeline.
While this marks an important advancement in how fairness assessment is integrated into the model development process, it is only designed for binary classifiers and reflects only a single sensitive class.
\\
In 2019, Cabrera et. al. presented FairVis~\cite{cabrera2019fairvis}, which is a rich visualization tool, where fairness metrics can be investigated for different intersectional subgroups.
FairVis additionally integrates methods for suggesting and discovering subgroups to investigate.
In the case of large label spaces, however, their proposed subgroup analysis is not applicable, as the attributes needed for such a subgroup analysis are often missing, for example when operating on unlabeled or sparsely labeled data.
Additionally, the visualizations that FairVis introduces are not designed for the large number of annotations that large label spaces contain.
\\
Similarly, the What-if Tool~\cite{wexler2019if} also supports the inspection of intersectional subgroups, and even allows for experimentation with individual data points to investigate counterfactuals.
However, it was also not designed with novel, correlation-based fairness metrics and large label spaces in mind.
\\
FairSight~\cite{ahn2019fairsight}, which is integrated into the decision making process with the proposed FairDM, is another visualization system that communicates conventional fairness metrics, which makes it incompatible with large label spaces.
\\
With Boxer~\cite{gleicher2020boxer}, different classifiers can be compared on the basis of selected data subsets.
While this concept translates well for other domains, where subgroups are to be compared, the proposed visualizations are not designed for the amount of labels that large-scale multi-label classifiers operate on.

While we found a considerable amount of work in the domain of visualizing ML-fairness, which indicates the importance of this field, we observed a gap in the literature when it comes to modern, large-scale problems.
The main concerns with existing work for this domain are that, first, the proposed visualizations do not work well with a large number of different labels, and second, that these visualizations are based on conventional fairness metrics, which cannot be used for large datasets without exhaustive ground truth.
To fill this gap, our research focuses on novel fairness metrics for large label spaces.

%-------------------------------------------------------------------------
\section{Fairness and Large Label Spaces}
In the following, we first provide definitions for the terminology used in this paper.
Then, we explain how fairness metrics can be used in the context of large label spaces, before illustrating the current practice of bias mitigation.

%-------
\subsection{Definitions}\label{subsec:definitions}
\begin{figure}[bt]
    \centering
    \includegraphics[width=0.6\linewidth]{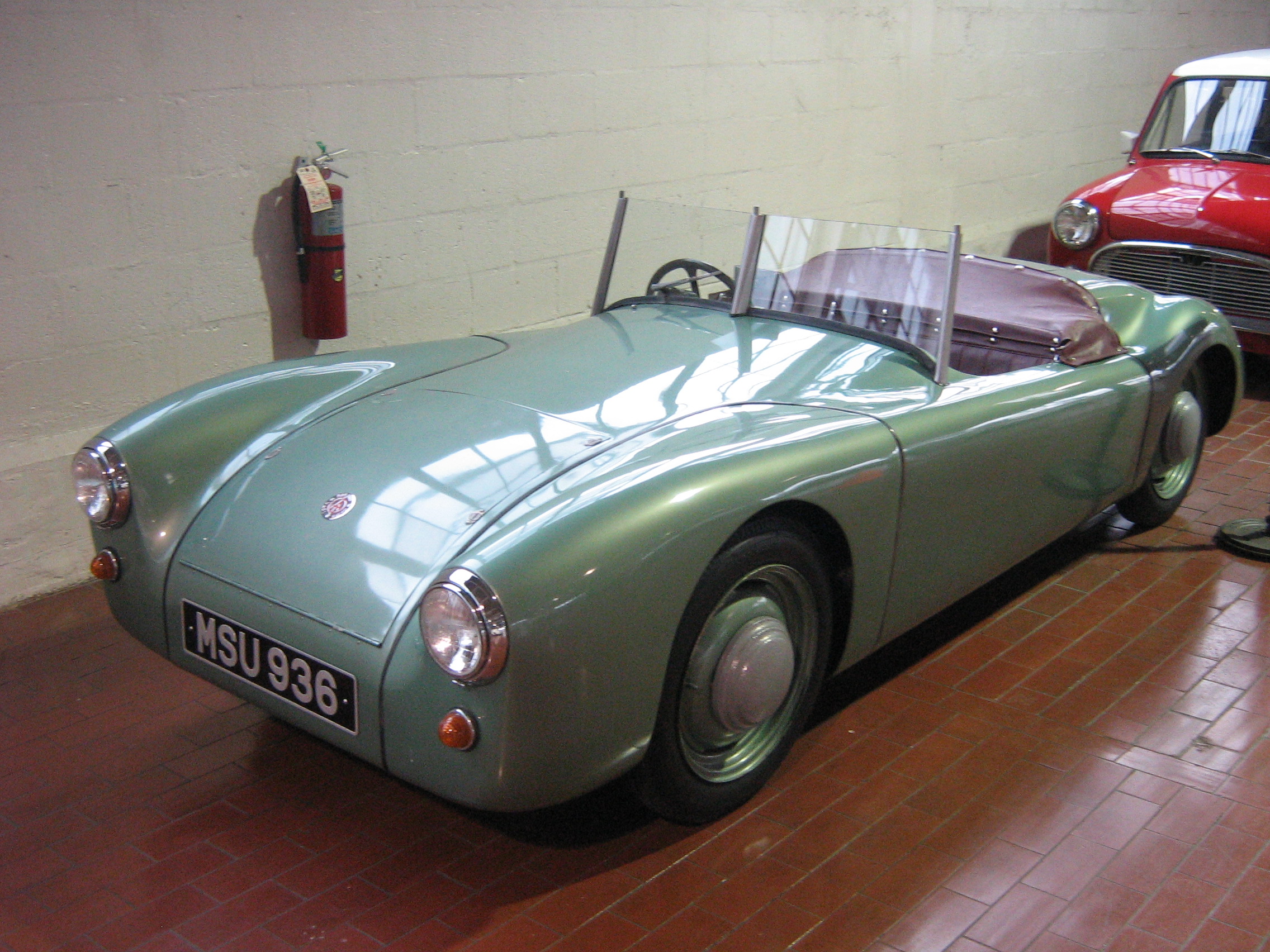}
    \caption{\label{fig:car} 
    Example for non-exhaustively labeled image.
    Positive labels in this image are \emph{Vintage Car, Race Car, Sports Car, Classic Car}, missing labels are, for example, \emph{Wheel, Light}, and \emph{Door}.
    The negative labels include \emph{Triumph tr3, Mg mga, Supercar}, and \emph{Austin-haley sprite}, negative labels such as \emph{Person, Tree}, and \emph{Ambulance} are missing.
    Image and labels from OIDv6~\cite{oid2020car}.
    }
\end{figure}

\textbf{Sensitive Attributes.}
Sensitive attributes are groups of labels or external annotations for which decision makers want to make sure there is no bias.
In most cases, sensitive attributes have 1 to at most 10s of directions which are compared to each other, such as \emph{male} and \emph{female} for gender, different skin colors for race, or geolocation for place of origin.
\\
\textbf{Large Label Spaces.}
We define large label spaces as settings, in which a dataset contains 1,000s of different labels or more.
Importantly, this refers only to the label-dimension, whereas the dimensionality of sensitive attributes for which bias investigations are conducted is mostly low.
\\
\textbf{Bias.}
Bias exists when a dataset or model is skewed towards one direction of a sensitive attribute in favor of the other with regard to a specific label.
Bias is not always of concern and thus does not always need to be mitigated.
Without being biased towards certain attributes, no learning would be possible (e.g. every tree classification will be biased towards the colors green and brown).
\\
\textbf{Problematic Bias.}
While bias is not necessarily something bad, problematic bias is.
Bias can become problematic when it discriminates between directions of a sensitive attribute for labels where this causes unwanted implications on the dataset or model use.
Whether bias is problematic remains an ethical question and thus depends on human evaluation.
Therefore, it cannot be solely determined by an algorithm.
\\
\textbf{Fairness Metrics.}
Fairness metrics can be used to measure bias in a dataset or model.
It is important to note that they can only expose where bias lies, not how much of a problem it poses.
\\
\textbf{Model/Dataset Fairness.}
The term fairness is used in this paper as a measure of how much problematic bias exists in a model or dataset.
As such, similar to the term problematic bias, it is human-defined.
A model or dataset is perfectly fair when it exposes no problematic bias, and unfair with regard to specific sensitive attributes and labels when those are problematically biased.
\\
\textbf{Exhaustive Ground Truth.}
Exhaustive ground truth is available for a dataset when we know about the presence or absence of every label for every data point.
An example for an image with non-exhaustive ground truth can be seen in~\autoref{fig:car}.
\\
\textbf{Bias Patterns.}
When operating on large label spaces, there often exist labels that are semantically related.
Thus, it can be important to spot patterns of similarly biased, related labels.

%-------
\subsection{Metrics}
Classical fairness metrics such as Statistical Parity~\cite{dwork2012fairness}, Equality of Opportunity~\cite{hardt2016equality}, and Predictive Parity~\cite{chouldechova2017fair} work well when exhaustive ground truth is readily available.
For non-exhaustive ground truth, these metrics would need to presume that the absence of a label implies the data point is negative for that label, which can be problematic.
For example, an image of a \emph{car} without a \emph{butterfly} label would imply the absence of a \emph{butterfly} without actually having this ground truth.
This could encode biases that ground-truth-dependent fairness metrics would by definition be unable to detect (in this case, that butterflies never appear in images with cars).
Thus, as ML models and datasets grow in size and label sparsity becomes more common, alternative ways to measure biases complementing ground-truth-dependent approaches must be used.
One approach uses correlation metrics as a proxy for bias in these settings~\cite{aka2021measuring}.
These metrics measure the co-occurrence of two features or predictions in a single data point (akin to the bag-of-words model in NLP~\cite{harris1985introduction}).
Using such metrics, one can calculate fairness scores for individual labels in large label spaces by comparing a label's co-occurrence rate between labels and sensitive attributes (e.g. compare the co-occurrence of \emph{basketball} with \emph{gender} directions).
Such metrics can be used to measure bias in both datasets and model predictions.
In the model context, one may wish to measure whether a classifier assigns a label at a high level relative to chance expectation for a direction of a sensitive attribute.
For datasets, we are interested in measuring the correlation between labels assigned to a data point and sensitive attribute directions, whereas label assignment can be through human annotators or other algorithmic approaches.
Correlation calculation can be done between different data labels or predictions of a multi-label classification model (e.g. \emph{scientist} and \emph{earring}).
Another option to calculate this correlation is between a data label or model prediction and another known attribute recoverable from metadata, such as geo-tags, time of capture, or even predictions from another neural network.
\\
Different correlation metrics can be used for assessing model fairness, e.g.~\cite{barocas2016big,dice1945measures,jaccard1912distribution,kendall1938new} and our visualization design has been developed to support fairness analysis based on any such metric.
However, for consistency and based on recent insights~\cite{aka2021measuring}, our experiments and evaluation were conducted entirely on the basis of normalized Pointwise Mutual Information (nPMI), as proposed by Gerlof Bouma~\cite{bouma2009normalized}:
\begin{equation}\label{eq:npmi}
i_n(x,y) = ln(\frac{p(x,y)}{p(x)p(y)}) / -ln(p(x,y))
\end{equation}

where $p(x)$ is the probability of a data point being associated with feature $x$, $p(y)$ is the probability of encountering feature $y$ for a data point, and $p(x,y)$ is the probability of a data point containing both features, $x$, and $y$.
For a labeled dataset or trained model, $x$ and $y$ would be predictions, labels, or known attributes of the data, as mentioned before.
This approach is rooted in information theory, whereby we are comparing the joint probability of the two labels, $p(x,y)$, to the expected co-occurrence rate under the assumption of independence (i.e. the product of the marginals, $p(x)p(y)$). 
The division by $-ln(p(x,y))$ normalizes values so that a value of +1 or -1 indicates that $x$ and $y$ always or never co-occur, respectively.
A value of 0 indicates $x$ and $Y$ co-occur at levels expected by chance alone.

%-------
\subsection{Current Practice}
Currently, lacking any usable visualization tool for investigating bias in large label spaces, decision makers typically look over spreadsheets that contain nPMI calculations for every label with respect to the dimensions of a selected sensitive attribute.
Hereby, labels are presented per row, while sensitive attribute dimensions are represented by the columns of this spreadsheet.
This approach has many major downsides, some of which are the lack of visual guidance, the lack of additional information, hard comparability of different runs or datasets, and limited filtering capabilities.
On the contrary, decision makers typically only sort the rows of these spreadsheets by one of the dimensions of the sensitive attribute, and then manually skim through the raw nPMI numbers without any visualizations or highlighting.
Thus, the need for a visualization tool with more functionality, better user guidance, and better visualization abstraction is big.

%-------------------------------------------------------------------------
\section{User Tasks}\label{sec:tasks}

To assess the fairness analysis needs of decision makers operating on large label spaces, we collaborated closely with two teams of expert practitioners throughout this project.
One of those teams is providing ML fairness infrastructure, while the other works on applications that utilize large label spaces, such that both of those teams are experts in our problem domain.
The fairness-assessment tasks that we extracted from these conversations are presented in the following.
\\
\textbf{\taskbox{T1} Discovering Problematic Bias.}
The first and most obvious requirement that the domain experts had was to be able to extract labels that expose problematic bias.
Typically, domain experts want to analyze bias with respect to certain sensitive attributes, such as gender, skin tone, or geospatial information.
Here, big correlation-gaps between different manifestations of those attributes, as well as extreme correlation values for individual labels, are of special interest to domain experts.
The main concern in these scenarios is discovering those problematic labels among thousands of labels in the dataset, which potentially are all biased towards a certain direction.
To be able to do this, domain experts additionally are interested in assessing whether a displayed correlation value is salient.
\\
\textbf{\taskbox{T2} Visually Analyzing Label Spaces.}
While domain experts can already compute correlation statistics for bias analysis, they miss a streamlined process for working through a labeled dataset.
To us, they mentioned that they would like to be able to systematically analyze the correlation calculations, and obtain actionable results after the process.
\\
\textbf{\taskbox{T3} Comparing Configurations.}
Another task that domain experts often have to perform is comparing different datasets or models.
Here, they are interested in whether, for instance, different labeling strategies, training procedures, or model architectures expose different biases.
Such comparative analysis approaches can inform decision makers about the difference between strategies to pick the ones that suffer from the least problematic bias.
\\
\textbf{\taskbox{T4} Finding Hidden Patterns.}
When talking to the domain experts, they also mentioned that they suspect bias patterns in the label-space, such as labels for certain professions being biased towards specific skin tones.
Those patterns, they suspected, would stem from related labels being biased similarly towards sensitive attributes.
Thus, domain experts need to be able to find out if semantically similar labels are biased towards the same sensitive attributes, or if the bias of a label is an exception in the label space.
This can inform decisions about how to change training processes or labeling strategies.
Therefore, discovering those hidden patterns is another task which these domain experts would like to be able to perform, but cannot with current approaches.

%-------------------------------------------------------------------------
\section{Visualization Design Challenges}

Based on our domain expert interviews and the tasks we extracted from these, as well as insights from recent research, we then extracted the following visualization design challenges.
When talking to the domain experts, we first asked them where they see problems currently in conducting their tasks (\taskbox{T1-4}).
Then, based on these insights, we compared these problems with regard to our setting of large label spaces to related work.
Challenges mark points where related work did not provide an adequate solution for our problem at hand and thus needs to be tackled differently by our visualization approach.
In this context, we extracted both technical and ethical challenges which illustrate the problems specific to fairness visualizations for large label spaces.

%-------
\subsection{Technical Challenges}
First, we elaborate on the technical visualization challenges that arise in large label spaces.
We define those as the challenges which are not specific to the fairness domain but are related to the nature of large label spaces.
\\
\textbf{\technicalbox{TC1} Label Space.}
The first visualization challenge comes from the large number of labels.
Data for large label spaces is collected only on the domains of~\emph{labels},~\emph{sensitive attributes}, and~\emph{metrics}.
While almost all of these data domains are relatively low-dimensional (1 to 10s), the dimensionality of the~\emph{labels} is in the thousands~\cite{imagenet_cvpr09,kuznetsova2018open}.
We, thus, face the challenge of operating on a large number of labels, all of which can expose problematic bias.
Therefore, our visual representation needs to provide an overview over all labels, but also provide means to discover labels of interest.
This problem has not been of major concern for more limited label spaces and associated visualization approaches.
\\
\textbf{\technicalbox{TC2} Configurations.}
Apart from a lot of data points, domain experts also want to compare different configurations (i.e. labeling strategies, datasets, models, or training runs).
This leads to another visualization dimension, making the data four-dimensional (labels, sensitive attributes, metrics, configurations).
Here, not only the biases for all the individual labels have to be accessible, but bias calculations should also be comparable across different configurations~\cite{munzner2014visualization}.
Again, most other visualization approaches mainly focused on single models or datasets, and none have considered such configurations in combination with the sheer amount of data we operate on.
\\
\textbf{\technicalbox{TC3} User Guidance.}
In the current setting, where domain experts look at large spreadsheets of correlation values, there is no clear entry point and no systematic process in place for analyzing potentially problematic bias.
Providing such clear entry points and a streamlined process towards discovering such biased labels is key in a visualization approach for domain experts~\cite{ceneda2016characterizing,ceneda2020guide}.
Additionally, progressing through the thousands of labels, essentially dividing the label space into problematic and non-problematic after an analysis session is something that visualizations need to support.

%-------
\subsection{Ethical Challenges}
Wherever domain experts need to make a fairness decision informed by visualization in large label spaces, said visualization may face the following ethical challenges.
\\
\textbf{\ethicalbox{EC1} Reliability.}
One aspect when working with correlation values is the trustworthiness of a bias calculation.
Correlation values are always dependant on the sample size they were calculated on~\cite{bouma2009normalized}.
If there are too few samples to base the metric calculation on, resulting metric values might not represent real data distributions well, and, thus, need to be double-checked.
A small number of samples for a feature or sensitive attribute can in fact be a warning sign in itself, indicating that either more data is needed, or the model is particularly bad at recognizing said feature.
Domain experts should, therefore, be able to access additional information about the correlation, to help them assess whether a correlation value is supported by representative data.
While visualizations based on classical fairness metrics do not face this challenge, it needs to be addressed by our approach in that such potential problems need to be visually communicated.
\\
\textbf{\ethicalbox{EC2} (Un)Problematic Bias.}
Another challenge is determining whether a high correlation value actually corresponds to problematic bias~\cite{eu2020altai,jobin2019global}.
Decision makers need to evaluate whether they think a certain bias reflects their intentions with the neural network, or if corrections need to be made~\cite{law2020designing}.
This subjective assessment of separating problematic from non-problematic bias is ultimately left to human evaluation in the data and modeling process.
Nonetheless, it should be supported by bias visualization approaches for large label spaces.
While this challenge also exists for other fairness visualization approaches, it needs to be tailored to our setting of comparing a large label space with specific sensitive attributes.
\\
\textbf{\ethicalbox{EC3} Bias Patterns.}
Another visualization design challenge is how to communicate hidden patterns and outliers in the data.
While individual labels can expose problematic bias, a whole set of semantically similar labels which are all biased towards a certain direction poses an even bigger fairness problem.
Here, similarity information for labels has to be considered and displayed in a way that lets domain experts reason about individual instances as well as bigger bias problems throughout the models and for different sensitive attributes~\cite{michaels1998cluster,munzner2014visualization}.
Again, this is especially important in our setting, where the label space is large, and there exist many related labels.

%-------------------------------------------------------------------------
\section{Design Guidelines}
Informed by the user tasks and design challenges we presented in the previous sections as well as visualization research, we propose the following visualization design guidelines for investigating bias in large label spaces.
\\
\textbf{\guidebox{G1} Visual Guidance.}
Any visual analytics approach should provide systematic guidance for its users~\cite{ceneda2016characterizing,ceneda2020guide}.
This guideline just gets more important when dealing with large data, such as in detecting problematic bias in large label spaces.
\\
Within this problem domain, decision makers are mostly interested in very specific correlations.
These can be extreme correlations for one direction of a sensitive attribute or correlation-differences between sensitive attribute directions (e.g. samples highly biased towards one gender).
Thus, users should be able to specify such attributes as a clear entry point.
Then, based on this selection, they should be informed about the distribution of correlation values for the selected attribute, thus giving them information on which bias range to look at first.
\\
As we are operating on high-dimensional data, we argue that users should be able to inspect just the subset of labels they are interested in (e.g. all labels that are correlating with female much more than male).
In this regard, we are following Andy Kirk's guidance: \emph{[...] filter out the noise from the signals, identifying the most valuable, most striking, or most relevant dimensions of the subject matter in question}~\cite{kirk2012data}.
To enable a focused inspection of such data subsets, we argue that after selecting a sensitive attribute direction of interest, domain experts should be able to filter by the correlation value for said sensitive attribute direction.
\\
To support this analysis, the ability to make progress through the dataset is essential.
Here users should be able to confirm or reject potentially problematic bias, thus cleaning up the data for actionable interventions.
Drawing inspiration from other correction approaches in the machine learning domain~\cite{bauerle2020classifier,xiang2019interactive}, we propose to enable marking of problematic labels, and removal of unproblematic ones from the visualizations.
\\
Using these filtering approaches, which are applied from coarse (sensitive attribute) to fine (correlation values), 
and providing useful overview visualizations throughout, we follow Shneiderman's Mantra~\cite{shneiderman2003eyes} of \emph{providing an overview first, then being able to zoom and filter}.
In addition to that, we provide means to progress through the data, altogether proposing a streamlined way of navigating these high-dimensional datasets.
\\
\textbf{\guidebox{G2} Comparability.}
We argue that model designers can make a decision on whether a specific dataset, model architecture, or training procedure outperforms another only with the help of visual comparison.
Thus, making these configurations comparable in the visualization is another important design guideline we propose for bias identification in large label spaces.
This includes comparing bias indicators between models or datasets for individual labels, as well as looking at the bias-distribution across different models or datasets.
Here, we can draw inspiration from Tamara Munzner, who states: \emph{the capability of inspecting a single target in detail is often necessary, but not sufficient, for comparison}~\cite{munzner2014visualization}.
We, thus, propose to visualize both the distribution of correlation values and values for individual labels to enable a visual comparison of different configurations.
\\
\textbf{\guidebox{G3} Bias Context.}
To assess the reliability of a bias indication for a specific label, domain experts have to look into the mathematical credibility of the observed metric~\cite{bouma2009normalized}.
Additionally, as mentioned earlier, marking exposed bias as problematic is a subjective decision~\cite{eu2020altai,law2020designing,zhang2020joint}.
To assess whether detected bias actually poses a problem, domain experts have to be able to relate the value to the label it was calculated for.
Then, they have to bring the biased label and metric it is biased towards back into the application context to make a decision.
\\
We thus propose to show both contextual information about the correlation calculation and the label context throughout any visualization for identifying such bias.
For the correlation calculation, we propose to display information about the number of samples on which this correlation value was computed in addition to the correlation value itself, whereas for label context, we advise to always display the combination of the original label information, the sensitive attribute direction, and the model or dataset context.
\\
\textbf{\guidebox{G4} Pattern Discovery.}
Our expert interviews and experiments exposed that, in large label spaces, domain experts need to be able to spot systematically biased label groups and differentiate these from individual labels that expose problematic bias.
Thus, we propose to visualize patterns and outliers in the bias-distribution of the data.
We argue that visualizations for large label spaces should incorporate and transform label similarity information into visualizations that expose biases within the context of label similarity.
This approach is similar to insights on semantic information extraction from the domain of natural language processing~\cite{bolukbasi2016man}.

%-------------------------------------------------------------------------
\section{Implementation}

In the following, we present the implementation of our visualization design, which is based on the design guidelines introduced in the previous section.
We open-sourced our implementation as a TensorBoard plugin and our approach can be integrated directly into classical model or data lifecycles, as shown in~\autoref{fig:process}.
\\
In this paper, we focus on the visual inspection step of this process, which is separate from the metric calculation step.
While an interactive selection of sensitive attributes would be compelling for smaller datasets, our approach is designed to work on large datasets $D$ (up to billions of samples), and label spaces $N$ ($>10,000$ labels).
The complexity of accumulating count values for one sensitive attribute direction is $O(D*N)$, which is computationally not feasible in an interactive way for such numbers.
Additionally, in most cases, decision makers are interested in specific sensitive attributes anyway.
Thus, we calculate nPMI scores for selected sensitive attributes offline.
\\
We also chose not to display source data for multiple reasons.
First, source data contains sensitive information in many cases, which cannot be shared broadly.
However, we want our bias analysis to be something to iterate on collaboratively and discuss in meetings.
More importantly, though, our datasets are large, and each potential bias is associated with many data samples, which cannot be provided interactively alongside the visualization.
At the same time, if we were to reduce the data size through aggressive sampling, we would not know if the sampled data well reflects the reason behind detected bias.
On the contrary, such sampling could itself introduce unwanted bias in the analysis process.
Thus, our bias detection approach does not include such examples.
This keeps our visualizations and gained insights shareable, objective, and manageable.

%-------
\subsection{Guidance Visualization}\label{subsec:guidancevis}

\begin{figure*}[tb]
    \centering
    \includegraphics[width=1\linewidth]{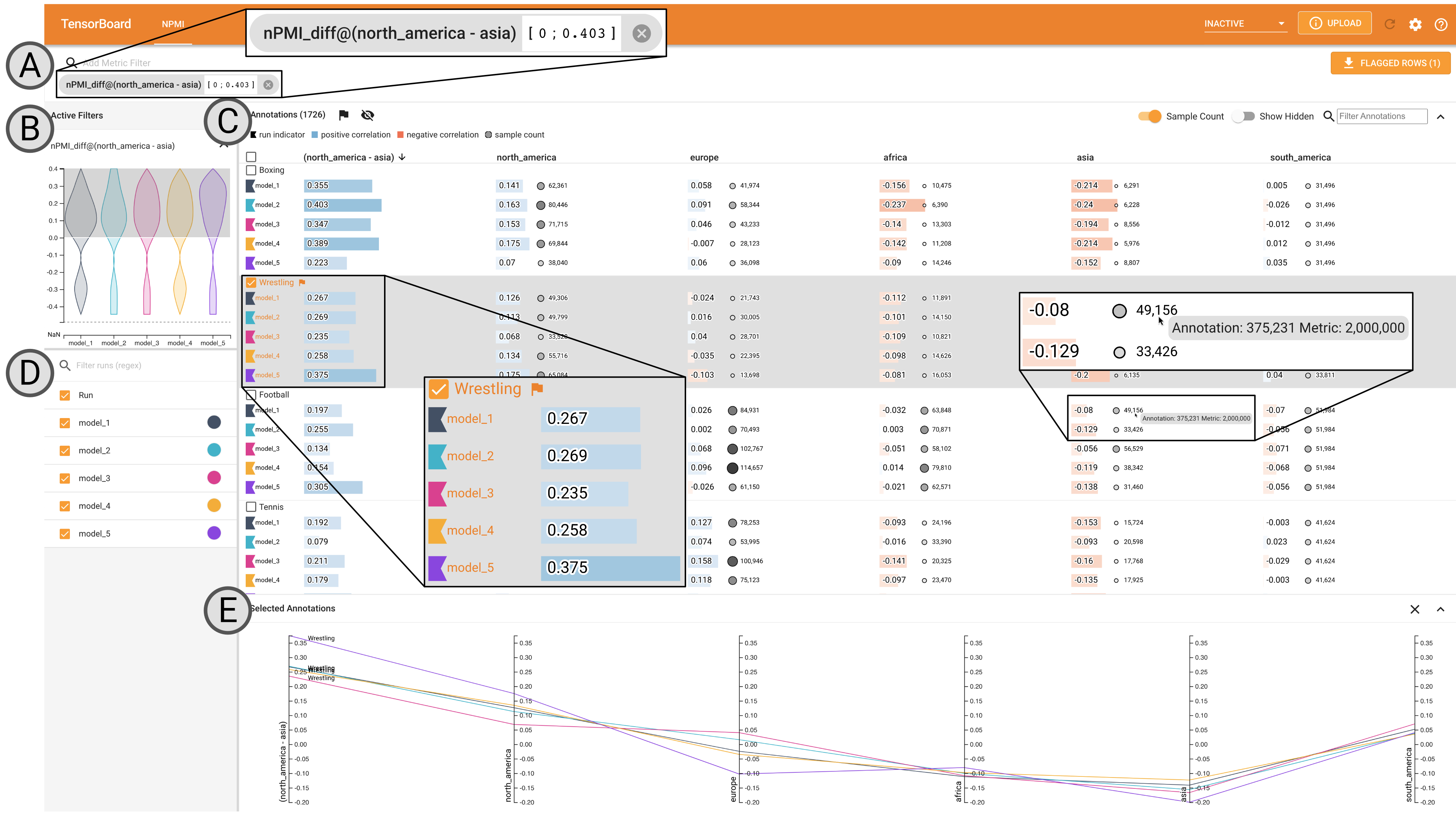}
    \caption{\label{fig:application} 
    Implementation of our techniques as a TensorBoard plugin applied to synthetic data.
    Annotations are filtered by a correlation metric \figurebox{A}, in this case, the nPMI difference between male and female.
    In \figurebox{B}, one can see the distribution of correlation values for the two inspected models, and brush to select a correlation-range to be displayed.
    In \figurebox{C}, the annotations which are filtered by the aforementioned filters are shown with their according correlation values and the number of data points a correlation calculation was based on.
    In \figurebox{D}, different models or datasets can be selected for display.
    In \figurebox{E}, a parallel coordinates visualization of selected annotations is shown, helping to visually assess where correlation differences come from, and where models differ.
    }
\end{figure*}

\begin{figure*}[tb]
    \centering
    \includegraphics[width=1\linewidth]{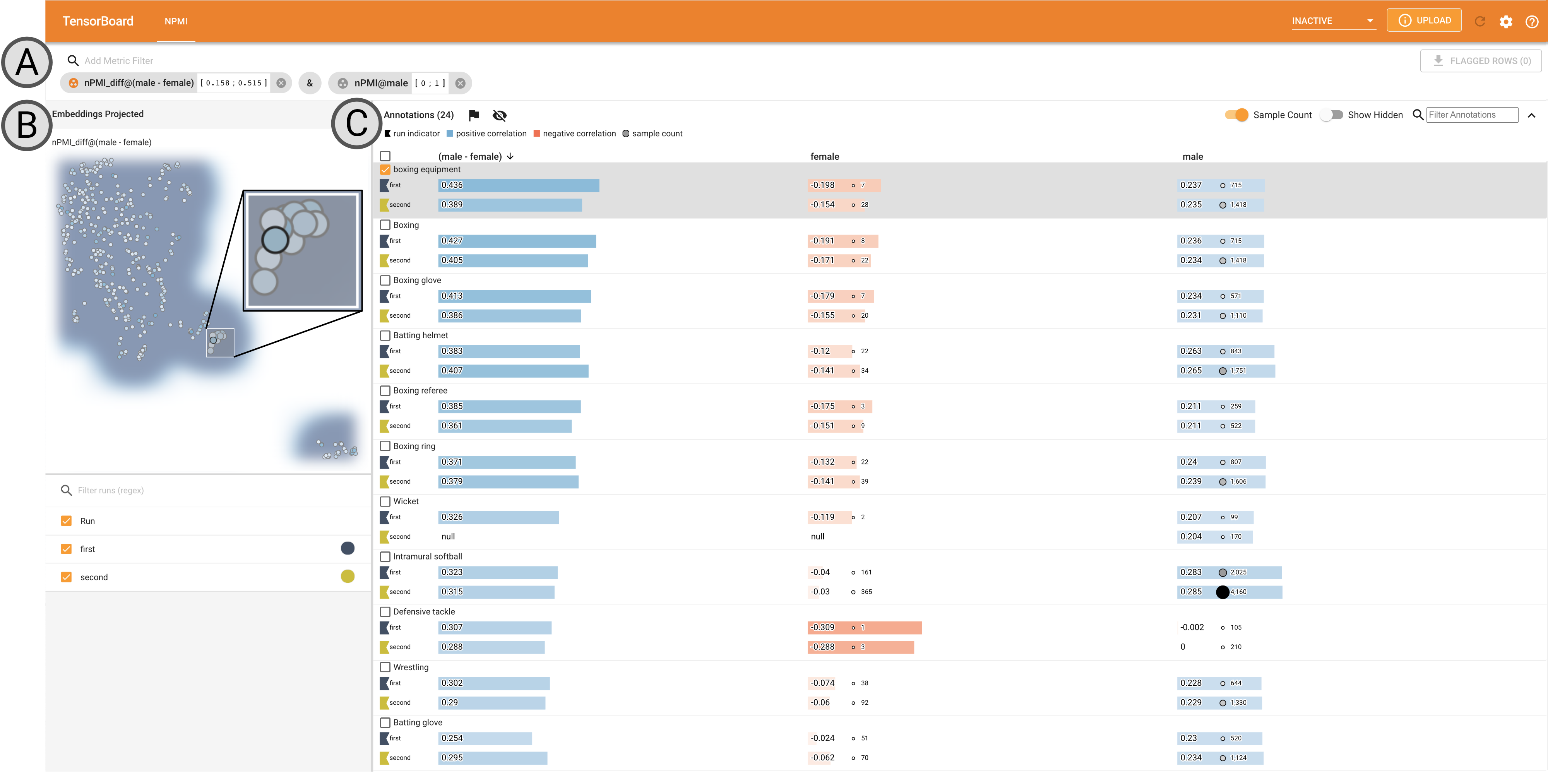}
    \caption{\label{fig:embeddings} 
    Embeddings view of annotations for selected metric.
    Visualization based on OID data.
    Here, two filters are defined, leaving only samples that are correlated much more with male than female, and have high correlation values with male \figurebox{A}.
    \figurebox{B} shows a UMAP projection of 25-dimensional embeddings of the annotations that are not filtered by the aforementioned metric filters.
    Here, annotations that are close are likely semantically similar, as in this case, a cluster of boxing-related annotations forms.
    This embedding view can be used to filter the annotation view in \figurebox{C}, allowing for embedding-based annotation inspection.
    In \figurebox{C}, one can also see how flagged annotations are highlighted in the annotation visualization.
    Additionally, annotations that are selected in \figurebox{C} are drawn with a black outline in \figurebox{B}.
    }
\end{figure*}

To support \guidebox{G1} with being able to select a data subset of interest, we provide correlation-based filtering methods.
This serves as a natural entry point to the visual analysis~\technicalbox{TC3}.
Using our filters, one can select a sensitive attribute direction and define a range of correlation values for it, effectively reducing the number of labels at display~\technicalbox{TC1}.
For instance, in~\autoref{fig:application} \figurebox{A}, the data was filtered by the nPMI difference between \emph{north\_america} and \emph{asia}, whereas only correlation values between $0$ and $0.403$ have been selected.
\\
We show a violin plot that displays the distribution of correlation values for each active metric filter, whereas multiple models or datasets are shown next to each other (~\autoref{fig:application} \figurebox{B}).
Here, brushing can be used to adjust the filter, which manifests in the annotations that are displayed in~\autoref{fig:application} \figurebox{C}.
The same behavior can be achieved by modifying the filter limits of~\autoref{fig:application} \figurebox{A}, which allow for a more exact filter specification.
With this visualization approach, we follow the \emph{Guidelines for using multiple views in information visualization} and implement brushing and linking~\cite{wang2000guidelines}.
Filters can also be combined, as shown in~\autoref{fig:embeddings} \figurebox{A}, such that even more fine-grained data subsets can be inspected.
\\
For the main visualization of the bias calculations (\autoref{fig:application} \figurebox{C}), we employ a visual encoding similar to the well-known spreadsheets that decision makers currently use.
While this places them in a familiar environment, we have made several visualization adjustments to improve user guidance and bias discoverability.
To guide the user's attention within this visualization, annotations are automatically sorted by their bias value when selecting a metric filter, whereas sorting can be changed by the user.
This way, users can vertically scan the annotations visualization to compare different annotations with respect to a sensitive attribute direction, as the bars align vertically for each sensitive attribute direction~\cite{waldner2019comparison,l2020comparative} while preserving the common vertical scrolling direction for the list of attributes.
We additionally use color-coded bars for visualizing the correlation values, following Mackinlay's ranking~\cite{mackinlay1986automating}, where color ranks just behind position for nominal data.
However, we employ position for the different metrics and labels to form a tabular display.
The length of these bars signals how strongly they correlate with a sensitive attribute direction, whereas the color is based on the direction of said correlation.
Here, positive correlations are visualized using a blue color scale, while negative ones are encoded using a red color scale.
\\
If space is sufficient, we always display all sensitive attribute directions as columns (\autoref{fig:application} \figurebox{C}).
Metric difference calculations are only displayed once the user adds a filter for such a metric, as users are only interested in one of those in most cases.
In the rare case that there is not enough display space for all attribute directions, we propose to fall back to only displaying directions with active filters to maintain scalability.
This way, users can investigate any number of sensitive attributes they have bias calculations for, showing all directions when possible, and only directions of interest when space is limited.
\\
Apart from a clear entry point and a data overview, users benefit from being able to make and see progress when working through these datasets~\cite{ceneda2020guide}.
Thus, we implemented both the removal of annotations that are viewed as non-critical and the flagging of annotations that expose problematic bias.
In our annotations view (\autoref{fig:application} \figurebox{C}), users can select individual or multiple annotations.
These selected annotations, which are displayed for further assessment in the parallel coordinates view in~\autoref{fig:application} \figurebox{E}, can then be hidden from all visualizations by clicking the \scalerel*{\includegraphics{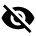}}{)} icon above the annotations visualization.
This removes annotations from all visualizations, and thus allows systematic progression.
To get back to hidden annotations, users can toggle between showing and removing annotations that are hidden.
Additionally, to mark annotations that expose problematic bias, selected annotations can be flagged using the \scalerel*{\includegraphics{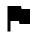}}{)} icon above the annotations visualization, which highlights these annotations (see \autoref{fig:application} \figurebox{C}).
Such flagged annotations can then be downloaded from within the implementation to generate actionable reports, altogether realizing a systematic data analysis process.
\\
In summary, with our visualization design, we first provide an overview through our violin plots.
Then, we allow filtering of the data to be investigated using our metric filters, before obtaining details on demand when looking at and selecting individual annotations~\cite{shneiderman2003eyes}. 
Finally, users can act upon their findings by systematically processing the dataset.

%-------
\subsection{Configuration Visualization}
To compare and evaluate multiple models or datasets~\technicalbox{TC2}, our visualizations are designed to display such configurations concurrently~\guidebox{G2}.
Typically, it is not a common use case to compare large numbers ($>5$) of configurations.
However, we still employ a run selection, which is similar to other TensorBoard plugins, and can be seen in~\autoref{fig:application} \figurebox{D}.
Thus, if many configurations are loaded, users can always select the subset they are interested in for keeping the visual comparison focused.
Our visualization design adapts TensorBoard's concept of assigning colors to individual runs, which is supported by perception research~\cite{christ1975review,mackinlay1986automating}. 
Here, the color value for a configuration matches the color values of runs from other TensorBoard plugins, which might be used in combination with our correlation analysis.
\autoref{fig:application} shows how our visualizations (\figurebox{C-E}) scale to multiple such configurations.
\\
Our filtering works across configurations, and only filters out annotations where none of the active configurations meets the filter criteria.
To support such filtering decisions, our metric overview visualization as shown in~\autoref{fig:application} \figurebox{B} displays correlation distributions for all active configurations simultaneously. 
This can be used to compare the overall bias tendencies for these different configurations.
We draw the violin plots for this correlation overview next to each other to prevent overdraw and to be able to convey all the correlation information for each model through our configuration-based color-coding.
\\
To compare biases between configurations, we display the correlation values for all configurations per annotation view below each other (\autoref{fig:application} \figurebox{C}).
Our bar chart-like visualization is designed to support this task, as users can simply vertically scan different configurations~\cite{waldner2019comparison,cleveland1984graphical} per annotation.
Directly comparing different metrics for sensitive attribute directions in our annotations visualization can be harder than comparing annotations or configurations, as bars for these metrics are not vertically stacked.
It can, however, be interesting to see how different configurations perform across those sensitive attribute directions.
Therefore, we also plot these metrics in our parallel coordinates view for selected annotations (\autoref{fig:application} \figurebox{E}).

%-------
\subsection{Context Visualization}\label{subsec:contextvis}
To address \ethicalbox{EC1}, we propose to always present the number of samples for a correlation calculation alongside the actual bias direction of a data point \guidebox{G3}.
This can be seen in~\autoref{fig:application} \figurebox{C}, where along with the bars that represent correlation values, we also show dots which are scaled and colored by the number of data points that a specific correlation value was calculated on.
Here, we use a white to black color scale, drawing the user's attention to the samples that are supported by the most dataset examples~\cite{christ1975review,mackinlay1986automating}.
Additionally, we show absolute numbers, and on hover, the numbers of how many data points match the annotation and metric for that configuration in total, if even more detail is required.
\\
The decision of whether a certain biased annotation poses a problem~\ethicalbox{EC2}, needs to be supported by the visualization design through embedding the correlation information with the label data~\guidebox{G3}.
To support decision makers in this process, our visualizations always communicate which annotation, metric, and configuration a bias belongs to (\autoref{fig:application} \figurebox{C}).
To keep the user informed about the label of investigation, we employ a tabular format for displaying label information, using vertical position for the labels at display, and horizontal positions for the metrics of interest, which domain experts are familiar with based on their previous use of spreadsheets.
Within each row of this visualization, we display the different configurations for each metric.
For all of these variables, we follow the guidelines of perceptual rankings for visualizing nominal data and employ a positional encoding~\cite{mackinlay1986automating}.
\\
Altogether, we provide insight into how specific correlation values were calculated.
Based on that information, domain experts can reason about the reliability of a correlation value \ethicalbox{EC1}.
Users can also connect bias with the annotation at inspection, and reflect upon their application domain to make an informed decision about whether a bias that has been exposed by the correlation calculation and visualized using our approach is actually of problematic nature \ethicalbox{EC2}.

%-------
\subsection{Pattern Visualization}
Biased annotations often only surface part of the underlying problem~\ethicalbox{EC3}.
We argue that visualizations of semantically similar annotations can expose bigger underlying bias problems~\guidebox{G4}.
\\
To support domain experts in their investigation of such \emph{bias clusters}, we provide an embedding view, as shown in~\autoref{fig:embeddings} \figurebox{B}.
This view can be accessed per metric by selecting the embedding icon on the left of a filter chip~\autoref{fig:embeddings} \figurebox{A}.
To realize such an embedding visualization, users can provide additional embedding information for the annotations in a dataset, which we project down to two dimensions using UMAP~\cite{mcinnes2018umap}.
This projection is done on the basis of the filtered data subset, and thus reflects only the data points the user is interested in at the moment.
For our experiments and figures in this paper, embeddings for all labels were obtained using \emph{Glove} word embeddings~\cite{pennington2014glove}, however, any embedding tool can be used with our visualizations.
\\
Our embedding visualization shows individual annotations as points, whereas the color of these points reflects the correlation direction, just as the bars we show in the annotation visualization.
Since projections can lead to overdraw, especially if many annotations pass the selected metric filters, we additionally provide a heatmap-based visualization behind this projection.
Here, where we use Gaussian kernels to spread the correlation value distribution across the canvas.
This functions as a visualization of the overall correlation distribution, helping decision makers to discover areas in the embedding view that expose similar biases.
\\
When users see individual annotations they want to inspect in the embedding, they can simply select those from the annotations visualization.
In the embedding visualization, such selected points are drawn with a black outline.
For the reverse direction, to make observations based on these embeddings, users can select points in this embedding space.
This selection is reflected as another filter in the annotations visualization.

Altogether, we mainly rely on well-known and time-tested visualization techniques, which we assembled specifically for use in the context of evaluating large label spaces on fairness concerns.
The intention behind this is to reflect, but greatly enhance the environment that users already know how to interact with, namely spreadsheets, to eliminate the need for visualization onboarding before being able to work with the presented techniques, and to thus enable users to make confident, well-informed decisions.
Nonetheless, our visualization design still simplifies the bias analysis in large label spaces greatly by adding introspection functionality that was not accessible before, such as advanced filtering methods, run comparison, compact visualization of both bias and additional information per label, and bias pattern visualization.

%-------------------------------------------------------------------------
\section{Usage Scenario}
Consider Susan, a model developer at a large software company.
Susan trained a model to tag photos of customers so that they can obtain albums based on what a photo contains.
She knows that a lot of customers that will use this feature are professional athletes from different regions all over the world.
Therefore, Susan wants to make sure that the model does not discriminate athletes of any heritage for certain types of sports \taskbox{T1}.
She has an anonymized dataset of her customers at hand, whereas each photo is labeled by the ethnic origin of the person depicted while doing a sportive activity.
Using this dataset and the predictions of the trained model, Susan uses nPMI as a correlation-based fairness metric between each label the model can assign and the sensitive attribute directions \emph{north\_america}, \emph{europe}, \emph{africa}, \emph{asia}, and \emph{south\_america}.
Then, she loads the data into our TensorBoard plugin to investigate potential bias.
\\
Susan knows of occasional findings in which Asian athletes were less likely to have their photos correctly sorted than North American athletes.
Susan wants to find out for which labels there is such a bias between Asian and North American athletes.
Therefore, she first selects the data subset she is interested in, using our metric filter on the difference between \emph{north\_america} and \emph{asia} (\autoref{fig:application} \figurebox{A}).
She sees the distribution of nPMI values for her model (\autoref{fig:application} \figurebox{B}), and limits the display of labels to those that are highly biased towards \emph{north\_america} \technicalbox{TC1\&3}\guidebox{G1}.
\\
Looking at the individual annotations \taskbox{T1}, by scanning them from top (most severe) to bottom (least severe), she notices two fighting-related labels that are highly biased.
After looking at the number of images on which these nPMI calculations are based on, and reflecting back on her customer domain as well as the labels and sensitive attribute she is looking at, she knows that this is indeed posing a fairness problem for her customers \ethicalbox{EC1\&2}\guidebox{G3}.
\\
She then loads up more models \taskbox{T3}, some older, some not yet released (\autoref{fig:application} \figurebox{D}), and compares the models' overall bias distribution, and bias with regard to the individual labels she found to be critical.
To be able to see the models across different sensitive attribute directions in even more detail, she uses the parallel coordinates plot of selected annotations.
She notices that, while the models expose slightly different bias, all are biased for these fighting-related labels she found and marks those labels as problematic \technicalbox{TC2}\guidebox{G2}.
\\
Susan now has the suspicion, that these are not the only fighting-related labels that expose bias \taskbox{T4}, and brings up the embedding visualization \autoref{fig:embeddings} \figurebox{B}.
After selecting one of the labels she found to be critical, she notices that it is embedded inside a cluster of labels.
Thus, she selects all data points within this cluster.
Now she sees that there is a bigger underlying problem, exposing that a lot of fighting-related labels are problematically biased towards male \ethicalbox{EC3}\guidebox{G4}.
Susan goes ahead and flags all of them.
\\
This way, Susan makes her way through the dataset, flagging labels that are exposing problematic bias, and hiding those that are not.
Finally, she can generate a report from these findings, which she brings to the next team-meeting.
Using Susan's report, her team finds out, that the data their models were trained on is biased already.
Thus, they go back to the data collection phase, and for the next iteration, investigate the dataset using the presented TensorBoard plugin before training a model on it.

%-------------------------------------------------------------------------
\section{Evaluation}
Throughout the process of developing this approach, we received continuous feedback from our partner teams.
We also ran a heuristic evaluation halfway through the project to improve upon our visualization design.
While the feedback we got during development as a result of the heuristic evaluation was informative and useful, after finalizing the visualization design, we also wanted to get feedback from potential users who had not seen our approach before. 
Thus, we conducted a virtual lab study, using the think aloud method to verify that our proposed approach in exploring big label spaces works as intended and can be used for the tasks we defined in~\Cref{sec:tasks}.

%-------
\subsection{Setup}
During the user study, we focused on nPMI as a fairness metric and used open-sourced predictions associated with OID~\cite{kuznetsova2018open}, which was randomly split into two subsets (60\%/40\%).
All predictions and visualizations were based on the whole OID label set (about 20,000 labels).
We calculated nPMI scores for all annotations with respect to the sensitive attribute directions \emph{male} and \emph{female}.
Then, we used our TensorBoard plugin to present the proposed visualization approach to our participants.
\\
We recruited six participants from a pool of potential users who had not seen our approach before.
Five of them were software engineers (2 female and 3 male) and one of them was a project manager (1 male), representing our target audience of model developers and decision makers.
Each study session took approximately 60 minutes.
Participants were given a quick introduction about how we use nPMI for identifying bias in large label spaces.
They then explored our implementation for one to two minutes to become familiar with it.
To evaluate our visualization design on the user tasks and challenges we extracted, we let our participants work on the following tasks:
\\
\\
\taskbox{T1} Problematic Bias 
\vspace{-0.6\topsep}
\begin{enumerate}
    \itemsep0em 
    \item Find biased labels
    \item Identify bias as (un)problematic
\end{enumerate}
\taskbox{T2} Visual Analysis of the Label Space
\vspace{-0.6\topsep}
\begin{enumerate}
    \itemsep0em 
    \item Filter and hide labels to reduce visualization complexity
    \item Generate exportable summaries of problematic labels
\end{enumerate}
\taskbox{T3} Comparing Configurations
\vspace{-0.6\topsep}
\begin{enumerate}
    \itemsep0em 
    \item Finding where model bias is similar
    \item Finding where model bias is different
\end{enumerate}
\taskbox{T4} Hidden Patterns
\vspace{-0.6\topsep}
\begin{enumerate}
    \itemsep0em 
    \item Identify semantically similar labels
    \item Reason about bias similarity/difference for similar labels
\end{enumerate}
For three of our participants, we said to focus on labels biased towards male and for the other three to focus on labels biased towards female. 
At the end of the session, we asked them how satisfied they were with their experience with the approach and if they had comments on our implementation and visualization design. 

%-------
\subsection{Results}
In the following, we are presenting the results from our user study.
\\
\noindent\textbf{Problematic Bias.}
For \taskbox{T1}, all participants were able to identify biased samples \technicalbox{TC1} using our filtering methods \guidebox{G1}.
While exploring individual labels, participants looked at the sample counts to assess the reliability of correlation values \ethicalbox{EC1}\guidebox{G3}.
Based on their insights and subjective assessment, they were able to separate problematic bias from unproblematic bias successfully \ethicalbox{EC2}.
Participants mentioned the most valuable feature for this task is the ease of data exploration.
P4: \emph{having the ease of the exploration would be the most valuable},
P5: \emph{before, we had lists like numbers of nPMI scores and someone would have to dig and manually sort or write like a processor or tool that would sort these nPMI scores}, and P5: \emph{Generally using the visualization is much easier, to be able to see patterns right away. It performs a lot of functions that the developer doesn't have to write themselves} \technicalbox{TC3}\guidebox{G1}.
\\
\noindent\textbf{Visual Analysis of the Label Space.}
Next \taskbox{T2}, our participants hid unproblematic labels and flagged problematic labels for further analysis \technicalbox{TC3}, \guidebox{G1}.
This allowed them to work their way through the data systematically, and clean up the label space one label after the other.
After these tasks, one of the participants already asked if the plugin was available for them to use as they already have a use case in relation to \taskbox{T1} and \taskbox{T2}.
In support of our highlighting and filtering approaches, participants mentioned: 
P3: \emph{We're trying to decide where we need to add human raters to labels. For something like that, I probably just go through the top couple hundred items} \taskbox{T1}\technicalbox{TC1} \emph{generate a list of candidate labels} \taskbox{T2}\technicalbox{TC3}\guidebox{G1} \emph{and then review those in a group} \ethicalbox{EC2}.
All participants were able to find labels that might be problematic \ethicalbox{EC2} and systematically hide or flag them \technicalbox{TC3}.
\\
\noindent\textbf{Comparing Configurations.}
For \taskbox{T3}, participants compared two different model configurations to see if those expose similar bias and if there is any label where the bias is different across models.
Again, participants were able to find both similar bias and different degrees of biases between the two models \technicalbox{TC2} using our visualization of these different configurations \guidebox{G2}.
\\
\noindent\textbf{Hidden Patterns.}
Finally, using our embedding visualization, participants looked at bias patterns \taskbox{T4}.
With our embeddings view \guidebox{G4}, our participants successfully identified semantically similar labels that were exposing similar bias in the data                \ethicalbox{EC3}.
This functionality of inspecting bias patterns that goes deeper than just one label in the dataset was what impressed our participants most about our approach:
P1: \emph{We literally can't do it at the moment [...] knowing that we've got a problem with a bunch of game and sports stuff being heavily biased towards male, like we know it's there but obviously we have no way of [...] properly understanding it}, and P1: \emph{Already looking at this you can say that well this is something that we just couldn't get access to before} (referring to embeddings view), or P4: \emph{having this tool to see what clusters are similar to one another and see the differences on the screen and everything, it would help a lot}, and P2: \emph{as far as I know there are no other tools to do this kind of analysis easily available}.
\\
\noindent\textbf{General Difficulties and Limitations.}
The main difficulties during our study came from the navigation within the plugin.
For example, it was not immediately clear to the users that they would need to click on the search bar \emph{Add Metric Filter} at the top to get the desired nMPI differences.
Also, while in our heuristic evaluation, users wanted us to keep the parallel coordinates view for detailed inspection of different labels, one participant of our user study mentioned:
P1: \emph{I haven't actually hundred percent figured out the usefulness of this section yet}, referring to the parallel coordinates graph, which reflects criticism also sometimes brought up in the visualization literature~\cite{henley2007evaluating,li2010judging}.
They additionally asked for an extension of our sorting feature, which currently supports sorting by a metric or by semantic similarity to a selected label, but not model differences.
P3: \emph{when we have an escalation of a new model to be able to just slap them on top of each other and compare them not only the distribution but also problem labels and generate a report from that would be amazing}.

%-------
\subsection{Discussion}
Overall, our approach proved to be helpful in solving the tasks that arise when assessing model fairness for the novel field of research on large label spaces.
As all participants were able to solve all \taskbox{tasks} that we presented, we conclude that our visualization \guidebox{guidelines} and implementation can be successfully used to tackle the \ethicalbox{ethical} and \technicalbox{technical} challenges we extracted.
This also indicates that our approach of using familiar visualization designs and well-known visualization paradigms works as intended.
Especially referring to our adoption of the tabular display of data which we enriched greatly with visual highlighting, additional information, and novel interaction concepts such as advanced filtering and flagging, users found exploring the data with our approach much easier than with the spreadsheets they had before, while still familiar enough to get started right away.
The comments related to our filtering methods and visualization guidance indicates that using our visualizations, domain experts could save a lot of time not having to look over the whole label space.
The flagging and export functionality was also a feature that decision makers apparently missed, as they mentioned that this would greatly improve their preparation for meetings and report generation.
Most well-received however, was our pattern visualization, which users mentioned to open up completely new ways of looking at bias in the large label spaces they operate on.
In their current practice, there is no process in place for doing such similarity-based analysis for large label spaces, and was, thus, not possible for the domain experts before.
Altogether, using our techniques, domain experts reported to be able to perform some tasks more efficiently than before, while others were just not possible without our approach.
The majority of the feedback we got from these participants was positive, where some of them even explicitly asked if they could include the plugin into the work of their team already, which indicates the need for such visualization approaches.

The main problems during our evaluation arose not from our proposed visualization design, but from navigation issues, which are neither a contribution nor the focus of this paper.
Here only points such as were where to click to get to a certain view (e.g. clicking the search bar to add a filter) were unclear.
Thus, while we will keep working to improve the navigation based on the user feedback we got, this does not influence the effectiveness of the techniques and insights we propose in this paper, which were received positively throughout.
Regarding the comment on the usability of the parallel coordinates plot, we suspect that this might be due to the fact that in our study, we only presented a two-directional sensitive attribute where visual comparison between these directions can be done even with only the tabular visualization.
However, we have seen in our heuristic evaluations that when comparing more directions of a sensitive attribute, the proposed parallel coordinates visualization can be helpful.
Additionally, the plot can always be hidden, such as in \autoref{fig:embeddings}.
Nonetheless, it might be worth looking into other ways of visualizing the difference between sensitive attribute directions across models.

%-------------------------------------------------------------------------
\section{Conclusion}
In this paper, we follow recent research and ethics guidelines of making bias analysis actionable through visualization.
What separates our approach from other existing fairness visualizations is that we focus on a different, modern problem domain, namely large label spaces, to which, so far, no approach has been targeted.
Through direct collaboration with domain experts, which led to a list of tasks that domain experts have to be able to perform, we were able to extract challenges and guidelines for visualization approaches in this domain.
These guidelines can help design visualizations to greatly improve the process, or open up new possibilities for assessing fairness in large label spaces.
We also provide an open-sourced implementation of the proposed techniques to allow for broad adoption and further improvements based on our ideas.
To evaluate our approach, we conducted a qualitative user study, where we tested the applicability and usefulness of our proposed visualization design in a realistic setting with domain experts.
This evaluation shows that our approach can be useful for performing the tasks that we extracted for such fairness problems with domain experts.

In the future, we want to add support for more metrics as we think that our approach can be helpful in many different scenarios that are related to what we discuss in this paper.
Additionally, we plan on improving the discoverability and adding further functionality to our implementation.
Another limitation of the current approach is the reliance on correctness for the sensitive attribute labels of the investigated data as otherwise, fairness metrics will not lead to accurate results.

\bibliographystyle{abbrv-doi}
\bibliography{VIPBLLS}
\end{document}